\documentclass[11pt]{article}

\usepackage[]{acl}
\usepackage{xcolor}

\usepackage{times}
\usepackage{latexsym}
\usepackage{graphicx}
\usepackage[T1]{fontenc}

\usepackage[utf8]{inputenc}

\usepackage{microtype}

\usepackage{xcolor}
\usepackage{csquotes}

\usepackage{caption}
\usepackage{subcaption}
\usepackage{booktabs}
\usepackage{threeparttable}
\usepackage{hyperref}
\def\hyphenateAndTtWholeString #1{\xHyphenate#1$\wholeString\unskip}
\def\xHyphenate#1#2\wholeString {\if#1$%
    \else\transform{#1}%
    \takeTheRest#2\ofTheString\fi}
\def\takeTheRest#1\ofTheString\fi
{\fi \xHyphenate#1\wholeString}
\def\transform#1{#1\hskip 0pt plus 1pt}
\def\urlx #1{\hyphenateAndTtWholeString{#1}}

\title{Unsupervised Mandarin-Cantonese Machine Translation }

\author{Megan Dare, Valentina Fajardo Diaz, Averie Ho Zoen So, Yifan Wang, Shibingfeng Zhang\\
Summer Semester Software Project 2022 \\
Language Science and Technology, Saarland University\\
{\{mdare,valenfd,averieso,yifwang,zhangshi}@coli.uni-saarland.de\}}

\usepackage{CJKutf8}

\begin{document}
\begin{CJK*}{UTF8}{bsmi}

\maketitle
\begin{abstract}
Advancements in unsupervised machine translation have enabled the development of machine translation systems that can translate between languages for which there is not an abundance of parallel data available. We explored unsupervised machine translation between Mandarin Chinese and Cantonese. Despite the vast number of native speakers of Cantonese, there is still no large-scale corpus for the language, due to the fact that Cantonese is primarily used for oral communication. The key contributions of our project include: 1. The creation of a new corpus containing approximately 1 million Cantonese sentences, and 2. A large-scale comparison across different model architectures, tokenization schemes, and embedding structures. Our best model trained with character-based tokenization and a Transformer architecture achieved a character-level BLEU of 25.1 when translating from Mandarin to Cantonese and of 24.4 when translating from Cantonese to Mandarin. In this paper we discuss our research process, experiments, and results.
\end{abstract}

\section{Introduction}
In recent years, neural machine translation has gained massive research interests. Most of these studies (e.g. \citealt{bahdanau2014neural, luong2015effective,wu2016google, vaswani2017attention}) focus on the construction of neural machine translation systems leveraging parallel bilingual corpora. Nevertheless, such an approach is not feasible for many language pairs due to the scarcity of resources for such pairs, as is the case for Cantonese and Mandarin. The study of automatic translation between these two languages faces the same problem: to the best of our knowledge, despite the vast number of native speakers of both languages, there is still no large-scale Mandarin-Cantonese parallel corpus. In addition, monolingual corpora for Cantonese are hard to collect as it is a low-resource language that is mainly used for only oral communication.

 Currently, only a few studies have been done on Cantonese-Mandarin translation, among which some compare various low-resource models for this language pair. However, these studies normally focus on a comparison between one or two model types. 
 Based on our motivation of implementing and training a Cantonese-Mandarin translation model and current state of research, we set our goal as building a robust model trained on a more diverse dataset, which can help improve communication between Cantonese and Mandarin speakers. Additionally, we seek to compare various model architectures, tokenization schemes, and embedding structures to conduct a comprehensive understanding on which settings may lead to the best performance for the Cantonese-Mandarin language pair.
 
After a close analysis of the current state of research and the available resources, we propose to develop a Cantonese-Mandarin machine translation system that is capable of conducting translation in both directions. The training of the system involves only Mandarin and Cantonese monolingual corpora collected from Wikipedia and various websites.

Our work also makes contributions to the Cantonese language NLP field by collecting Cantonese textual data and building a public large-scale monolingual corpus, which did not exist until now. In addition, considering the similarity between Cantonese and Mandarin, our translation system will provide a foundation for further development regarding machine translation tasks that center around language pairs composed of two similar languages.

\section{Background}
\subsection{Cantonese and Chinese: an overview}
Cantonese is one of the most widely spoken varieties of Chinese other than Mandarin Chinese \cite{matthews2013cantonese}. It is estimated to have more than 55 million native speakers, with large populations found in southern China provinces Guangdong and Guangxi, as well as regions including Hong Kong and Macau, it is also commonly spoken in overseas Cantonese communities in Singapore, Malaysia, North America and Australia as a result of emigration \cite{matthews2013cantonese}. 

While numerous NLP applications have been developed for Mandarin Chinese, little has been developed for Cantonese. One reason for this is the limited linguistic resources that have been collected for Cantonese. Primarily a spoken language and a non-standard variety, written Cantonese is not traditionally used or taught in schools. Instead, Cantonese speakers typically learn to read and write in standard Chinese through education, so there is no language barrier for Cantonese speakers when interacting with computer applications designed in standard Chinese. 

On the other hand, with the availability of the internet and the rise of social media, Cantonese is much more commonly used and written online in recent years, which can be seen as an indicator for a market in Cantonese NLP applications. 

It is important to note that this phenomenon might only be applicable to Hong Kong Cantonese, and not other variants such as the one in Guangdong province. More recent discussions about Cantonese, such as \citet{bauer2018cantonese}, make a point to distinguish between the Hong Kong Cantonese variant and the others, since the use of Cantonese is on the rise in Hong Kong, while declining in provinces within mainland China. Not only has this led to Hong Kong being named  \emph{``the Cantonese-speaking capital of the world"} \cite[p.64]{bolton2011language}, but also the rise of written Cantonese locally and subsequently, the Cantonese text data that are available online, which are of the Hong Kong variant of Cantonese. 

\subsection{Linguistic Differences between Cantonese and Mandarin}
Despite the common misconception that Chinese dialects share the same grammar, Cantonese and Mandarin are different at phonological, lexical and syntactic levels, and are not mutually intelligible \cite{matthews2013cantonese}. Some suggests it is more accurate describe Cantonese as a distinct language of the Chinese language family \cite{snow2004cantonese}. For the rest of this section, we describe some features that differ between Mandarin and Hong Kong Cantonese. 

\subsubsection{Writing Systems}
To anyone who can read Chinese, the most notable visual variation in written Chinese is the writing system - Traditional or Simplified Chinese. The two systems are equivalent to each other, and have one-to-one correspondence for each character. The following is some examples of traditional / simplified characters: ``open" \begin{CJK*}{UTF8}{bsmi}開\end{CJK*}/\begin{CJK*}{UTF8}{gbsn}开\end{CJK*}, ``talk" \begin{CJK*}{UTF8}{bsmi}話\end{CJK*}/\begin{CJK*}{UTF8}{gbsn}话\end{CJK*} and ``book" \begin{CJK*}{UTF8}{bsmi}書\end{CJK*}/\begin{CJK*}{UTF8}{gbsn}书\end{CJK*}. The usage of either system is primarily due to regional difference, with mainland China using the simplified system, while Hong Kong and Taiwan use the traditional system. 

\subsubsection{Lexical and Syntactic comparisons}
Vocabulary difference is the main barrier which prevents Mandarin speakers from understanding Cantonese \cite{snow2004cantonese}, it is also the aspect which is the most distinguishable between Cantonese and Mandarin. According to \citet{snow2004cantonese}, written Cantonese in formal domains can contain around 10-15\% Cantonese-only characters, while this percentage in informal domains can go up to 25-40\%.  Notably, the vocabulary that differ are some of the most frequent words, including many function words, as seen in Table \ref{table:1}.
\begin{CJK*}{UTF8}{bsmi}
\begin{table}[!ht]
\begin{tabular}{|p{3cm}|p{1.6cm}|p{1.6cm}|}
 \hline
 \textbf{Meaning} & \textbf{Cantonese} & \textbf{Mandarin} \\ 
 \hline
 possessive marker & 嘅 \textit{ge3} & 的 \textit{de}\\
 \hline
 perfective marker & 咗 \textit{zo2} & 了 \textit{le}\\
 \hline
 pronoun pluralizer & 哋 \textit{dei6} & 們 \textit{mén}\\
 \hline
 negator & 唔 \textit{m4} & 不 \textit{bù}\\
 \hline
 is (copula) & 係 \textit{hai6} & 是 \textit{shì}\\
 \hline
 this & 呢 \textit{ne1} & 這 \textit{zhè}\\
  \hline
\end{tabular}
\caption{Examples of lexical difference between Cantonese and Mandarin from \citet[p.49]{snow2004cantonese}. Cantonese romanizations follow the Jyutping system. }
\label{table:1}
\end{table}
\end{CJK*}
Syntactically, Cantonese and Mandarin are broadly similar but with some differences that are often overlooked \cite{matthews2013cantonese}. Some common differences are in terms of word order, including indirect object and comparative constructions \cite{snow2004cantonese}:

\begin{displayquote}
\underline{Indirect object construction:} \\
Cantonese: \\
我俾錢佢\textit{ngo5 bei2 cin4 keoi5} \\
(I + give + money + he)\\
Mandarin: \\
我給他錢 \textit{wó gěi tā qían}\\
(I + give + he + money)\\
`I give him money'\\
\\
\underline{Comparative construction:} \\
Cantonese: \\
我高過佢 \textit{ngo5 gou1 gwo3 keoi5} \\
(I + tall + more than + he)\\
Mandarin: \\
我比他高 \textit{wó bǐ tā gāo}\\
(I + compared to + he + tall)\\
`I'm taller than him.'

\end{displayquote}

\subsubsection{Challenges Unique to Cantonese NLP}
Firstly, there exists a certain degree of variability in written Cantonese since it was never standardised. As such, some words can be written with completely different characters yet have the same meanings and pronunciations. For example, ``like" can be written as 中意 or 鍾意 (read: zung1 ji3\footnote{romanizations according to the Jyutping system.}), ``still" can be written as 仲 or 重 (read: zung6) \cite{matthews2013cantonese}. Additionally, when some Cantonese words cannot be represented by existing Chinese characters, they could be written in a romanized form, such as the comparative (eg. ``-er" in ``cheaper") can be written with ``D", as well as a non-romanized form 啲 (read: di1) \cite{snow2004cantonese, matthews2013cantonese}. 

Secondly, code-switching to English is a common phenomena in Cantonese, which is not a feature in standard Chinese or Mandarin. Code-switching in Hong Kong Cantonese is mostly intrasentential (below clause level) \cite{li2000cantonese}, for example:

\begin{displayquote}
我哋今朝9點有個meeting。\\
ngo5 dei6 gam1 ziu1 gau2 dim2 jau5 go3 MEETING\\
`We have a meeting at 9am today.' 
\end{displayquote}

\section{Related Work}
\subsection{Unsupervised Machine Translation}
Unsupervised machine translation with no parallel data is a challenging task that has attracted many interests. The presence of cross-lingual embeddings \cite{mikolov2013exploiting, artetxe2016learning, artetxe2017learning, artetxe2018generalizing, artetxe2018robust, conneau2017word} provides prior knowledge for machine translation systems and makes it possible to train a machine translation model in an unsupervised way. \citet{artetxe2017unsupervised} and \citet{lample2017unsupervised} are the first attempts to explore the possibility of constructing a neural machine translation system using only monolingual corpora from both source and target languages. The proposed system is based on an encoder-decoder architecture with attention mechanism~\cite{bahdanau2014neural}, trained with a denoising auto-encoding task \cite{vincent2008extracting} and a back-translation task \cite{sennrich2015improving}. The encoder is shared by both the source and target languages, so that sentences from both languages can be mapped to a common latent space, while each language has its own decoder to reconstruct encoded sentences back into its own language space. Cross-lingual embeddings are leveraged as an initialization for the system, providing additional lexical level information. Such a structural property allows the translation model to be bi-directional, that is, the same model can be employed in both the L1-to-L2 translation task and the L2-to-L1 translation task.

This approach is extended in \citet{lample2018phrase} by applying a transformer model and using sub-word level tokenization methods. Attention-only structures provide higher model capacity, and sub-word level tokenization method Byte Pair Encoding (BPE) reduce the size of vocabulary and helps solving <UNK> problems in translation.  Additionally, they re-exploited the potential of statistical approaches in unsupervised machine translation tasks. A phrase-based machine translation model initialized with an automatically populated phrase table and language model is trained by iterative back-translation. Results of the experiment show that a statistical approach can reach similar performance or even outperform neural systems when the data is scarce, as the neural model tends to overfit the corpora, and thus does not generalize well. Together with \citet{singh2020unsupervised}, they show that unsupervised approaches can be used to construct machine translation systems for low-source languages (e.g., Urdu, Romanian, Manipuri).

In recent years, pre-trained language models have become popular due to their competitive ability of representing and generating natural languages learned from transfer learning on large-scale self-supervised datasets. Lample and Conneau \citet{lample2019cross} take their work one step further by pre-training both the encoder and decoder in their model using a cross-lingual language model (XLM). They then fine-tune the pre-trained model to an unsupervised neural machine translation model following the training process described in \citet{lample2018phrase}. The pre-training stage results in a sharp BLEU score increase over previous benchmarks for unsupervised machine translation.

Unsupervised machine translation methods are also applied in dialectal machine translation tasks, where the similarity and commonality between languages can be leveraged. \citet{FARHAN2020102181} uses common words between Arabic dialects as anchor points to steer projections of surrounding words between two dialects, creating a more accurate mapping between source and target words. In this way, they construct an unsupervised machine translation system with a BLEU score of 32.14, which is remarkably high compared with the highest BLEU score obtained in the supervised setting (48.25).

\subsection{Mandarin-Cantonese Machine Translation}

Due to the scarcity of available datasets, Cantonese language is always under-researched in NLP tasks. This issue is even more severe in machine translation tasks, which usually requires large amount of parallel data. For this reason, many researches on Cantonese-Mandarin machine translation are intended to collect more data or to fully exploit the limited data in a semi-supervised or unsupervised way.

Hei Yi Mak and Tan Lee \shortcite{yi2021low} construct a large-scale Cantonese-Mandarin parallel dataset by mining parallel sentences from Mandarin and Cantonese Wikipedia. They apply a similarity-based sentence alignment approach and use sentence pairs with high confidence score as parallel sentences. In this way, they end up with a parallel corpus of about 100,000 sentences. They also fine-tune a pre-trained language model using the collected data and obtain a competitive translation system that outperforms Baidu Fanyi, a commonly used translator in China. 

Concurrently, some efforts have been made to create unsupervised Cantonese-Mandarin translation systems. \cite{wan2020unsupervised} handles Cantonese-Mandarin translation as a dialect translation problem. which attempts to exploit the commonality between two language dialects. On the basis of \cite{lample2018phrase}'s transformer model, they make use of pivot-private embeddings and layer coordination to better utilize the similarity and difference between the two languages. Trained on two large monolingual datasets of 20 million colloquial sentences for each Mandarin and Cantonese, their model reaches an improvement of up to 12 BLEU score for Cantonese to Mandarin, and 5 BLEU from Mandarin to Cantonese compared to their baseline transformer model.

There have been other works relying on pre-trained cross-lingual language models (XLM). In \citet{wong2022mixed}, the authors initialize the encoder and decoder with XLM as described in \cite{lample2019cross}, while using pivot-private embeddings rather than cross-lingual embeddings. Using this enriched structure, they are able to achieve slight BLEU score improvements over previous XLM models.

\section{Corpus Construction}
While existing Cantonese corpora are scarce, and usually collected for linguistic purposes which is smaller in scale and of a specific demographic (eg. \citealt{wong2017quantitative,luke2015hong}), text data is available on the internet due to Cantonese being the common language used on social media. This also led to a rise in Cantonese writing in traditionally more formal domains such as advertisements, online news and subtitles. 

Therefore, we aim for the corpus to span across various domains for a comprehensive collection of modern Cantonese usage. Secondly, since standard Chinese is also commonly used among Cantonese speakers in online settings, in the data selection process, we aim to avoid sources which use standard Chinese. Lastly, in our pre-processing, we preserve some unique features in Cantonese such as code-switching in English. Detailed data statistics of the corpus is available on the Github repository. 

As we focus on collecting data for Cantonese, note that we simply use the Chinese Wikipedia for Mandarin data, since there is already a large amount of data available just from one source. 

\subsection{Data Collection}
\begin{figure}[t]
    \includegraphics[width=\columnwidth]{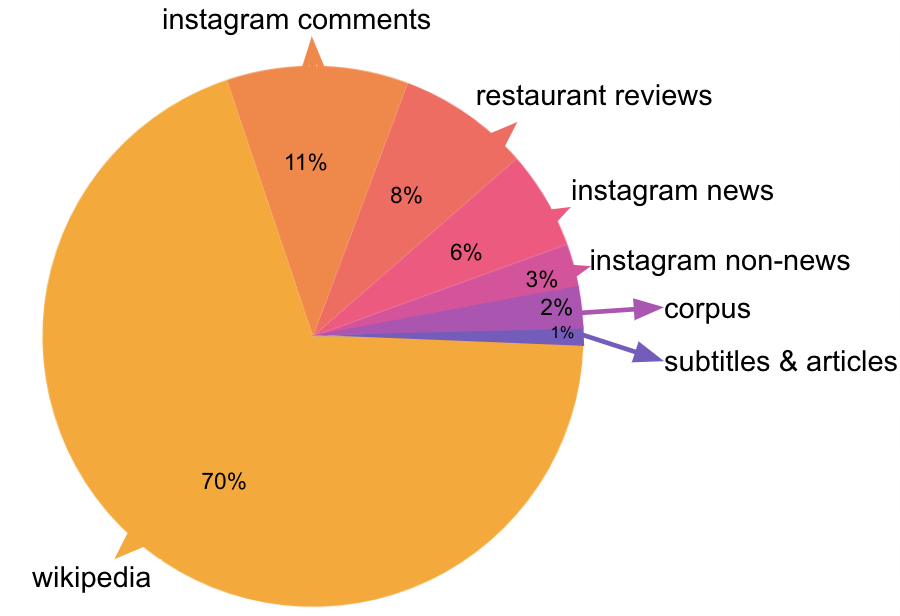}
    \caption{Distribution of data domain in the Cantonese training set (monolingual data only).}
    \label{fig:domain}
\end{figure}

The Cantonese data available from various sources on the internet are either readily downloadable (for Wikipedia, corpus and dictionary) or are scraped by us (for Instagram, subtitles and articles). Due to structural differences in the various websites, scraping functions are individually written for each of the three classes of sources. In general, the script moves recursively over the website domain and extracts any text in each web page. The scraping script is available on our GitHub repository. Figure \ref{fig:domain} shows the distribution in data domain of the Cantonese training dataset, which contains only monolingual data sources. 

\subsubsection{Monolingual Data}
\paragraph{Cantonese Wikipedia} The largest source of data available was Cantonese Wikipedia, which was downloaded from Wikimedia dump\footnote{https://dumps.wikimedia.org/zh\_yuewiki/20220601}, then pure text data is obtained with WikiExtractor \cite{Wikiextractor2015}. Cantonese Wikipedia amounts to 690k lines of text, making up 70\% of the Cantonese corpus overall. 

\paragraph{Corpus} As mentioned, there is a small number of open source Cantonese corpora collected for academic purposes, mainly transcribed from spoken Cantonese. Additionally, there is another corpus which contains scraped text data. Existing corpora add up to 95k lines of Cantonese text, with the majority coming from Openrice restraurant reviews (78k). 
\begin{itemize}
    \item openrice-senti\footnote{https://github.com/toastynews/openrice-senti}: scraped restaurant reviews from popular Hong Kong website OpenRice (\url{https://www.openrice.com/zh/hongkong}). 
    
    \item HK Cantonese Corpus\footnote{https://github.com/fcbond/hkcancor} \cite{wong2017quantitative}: manually transcribed oral conversations recorded between 1997-1998, includes spontaneous speech as well as radio programmes. 

    \item tatoeba\footnote{https://tatoeba.org/en}: a website which contains crowd-sourced sentences and their translations in many languages, including Cantonese. 
\end{itemize}

\paragraph{Instagram} Due to its popularity in Hong Kong, the domains from Instagram can be varied, ranging from blogs, advertisements, news and governmental organisations. We scrape posts and comments via \url{imginn.org} from 14 accounts, 5 of which are categorised as news, the others are categorised as non-news. Instagram comments make up the second largest source of Cantonese data with 108k lines (11\%), while Instagram news are 58k lines and Instagram non-news 30k lines. 

\paragraph{Subtitles} Cantonese YouTube\footnote{\urlx{https://docs.google.com/spreadsheets/d/1CmN8GPalrb45YFIPrWgh7GRYyoUhnizEOImY6kAW82w}} is a crowd-sourced compilation of youtube videos with spoken Cantonese subtitles. It is a voluntary effort from Cantonese learners, and each video is manually tagged with ``Written Cantonese" or ``Standard Written Chinese", which allows us to filter for only Cantonese videos. We are able to scrape directly from Youtube with the help of the Youtube Transcript API\footnote{https://github.com/jdepoix/youtube-transcript-api}. There are 1,620 lines. 

\paragraph{Articles} We scrape blog articles written by various authors in Cantonese from the freelancer platform \url{https://handstopmouthstop.com}. There are 6,531 lines from the website. 

\subsubsection{Parallel Data}
As the experiments described in the future sections are unsupervised, parallel data is not included in the training set. They are only used for the test set. 

\paragraph{Corpus} Cantonese-HK and Chinese-HK Universal Dependencies Treebank\footnote{https://github.com/UniversalDependencies/UD\_Cantonese-HK}\cite{luke2015hong}: manually transcribed and annotated film subtitles and legislative proceedings of Hong Kong, in both Cantonese and Mandarin. There are 1,004 parallel sentences from this corpus. 

\paragraph{Dictionary} Kaifangcidian\footnote{https://kaifangcidian.com/han/yue/} is an online Cantonese-Chinese dictionary which comes with parallel sentences for each lexical entry. There are 13,004 parallel sentences from the dictionary.

\paragraph{Subtitles} Kongjisubtitles \footnote{https://sites.google.com/view/lihkg-kongjisubtitles} is a Cantonese subtitle team that specialises in ``kongji"(meaning ``Hong Kong words" in romanized Cantonese) and focuses on subtitling Thai online series. Since some of the same videos also have Mandarin subtitles, we align them based on the timestamps of the videos. This amounts to 77,479 lines of parallel data. 

\subsection{Pre-processing}
Our data is scraped from different resources and inevitably contains noise. The following tools are leveraged for the pre-processing of collected data:

\paragraph{Sentence Cutter}
Sentence cutter cuts each text into sentences. The cutting points are punctuation marks such as 。.!? that defines the end of a sentence.

\paragraph{Mandarin-Cantonese Filter}
Due to the fact that most Cantonese speakers are also native in Mandarin, Mandarin text is normally present in Cantonese data scraped from social media. Mandarin-Cantonese Filter aims to determine whether a sentence is written in Mandarin or Cantonese by calculating the number of language-specific characters. This tool is involved only in the pre-processing of Cantonese data.

Cantonese-specific characters are: 咗, 唔, 係, 喺, 啦, 嘅, 既, 咁, 佢, 哋, 冇, 仲, 嘢, 乜, 噉, 咪, 咩, 俾, 呢, 嚟, 黎, 啫, 喂, 喇, 喎, 睇

Mandarin-specific characters are: 是, 的, 他, 她, 沒, 也, 看, 說, 在,\begin{CJK*}{UTF8}{gbsn}说
\end{CJK*}
\paragraph{Foreign Text Filter}
Text written in foreign languages such as Russian, Japanese and Korean abounds in collected data. Foreign Text Filter serves to filter out all sentences that are not written in Chinese characters. If the Chinese characters contributes to less than 5\% of sentence's total length, the sentence is removed. 

\paragraph{url, emoji, hashtag Remover}
This tool serves to remove url, emoji, and hashtag from sentence using regular expression.

\paragraph{Jieba Tokenizer}
Jieba~\footnote{https://github.com/fxsjy/jieba} is a Mandarin NLP library. In our project, we used Jieba tokenizer to pre-process our Mandarin data.

\paragraph{PyCantonese Tokenizer}
PyCantonese~\footnote{https://pycantonese.org/} is a Cantonese NLP library. In our project, we used PyCantonese tokenizer to pre-process our Cantonese data.

We did not include any Mandarin data from social media in our dataset, considering that data scraped from social media is always full of noises and Mandarin data from Wikipedia is already abundant for our task. We included Cantonese data scraped from social media since Cantonese data from Wikipedia is not sufficient.

\subsubsection{Overall Data Statistics}
\begin{figure}[t]
     \centering
     \begin{subfigure}[h]{0.48\textwidth}
        \centering
        \includegraphics[width=\textwidth]{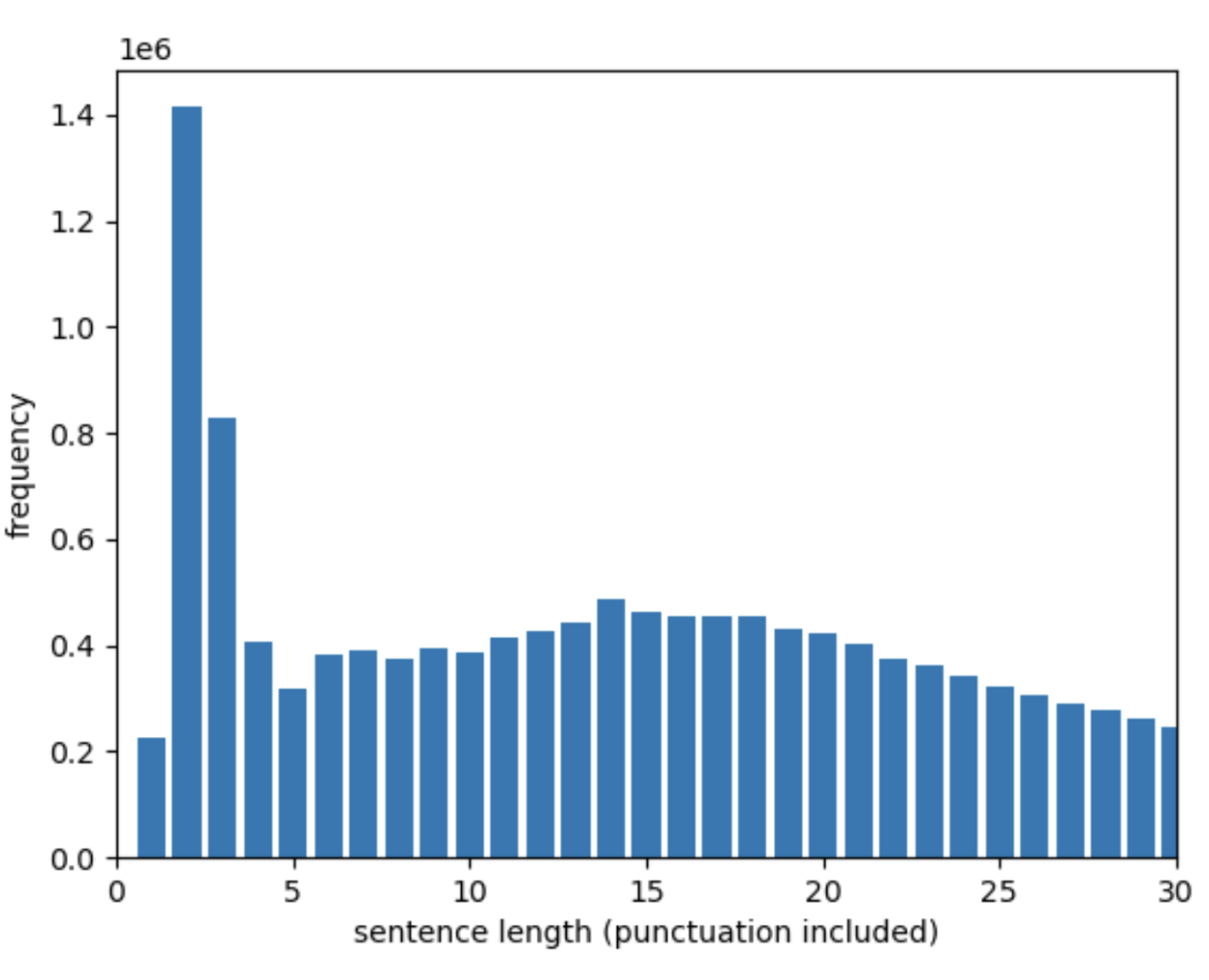}
        \caption{Mandarin corpus}
        \label{fig:mandarin_len}
    \end{subfigure}
    \hfill
    \centering
     \begin{subfigure}[h]{0.49\textwidth}
        \centering
        \includegraphics[width=\textwidth]{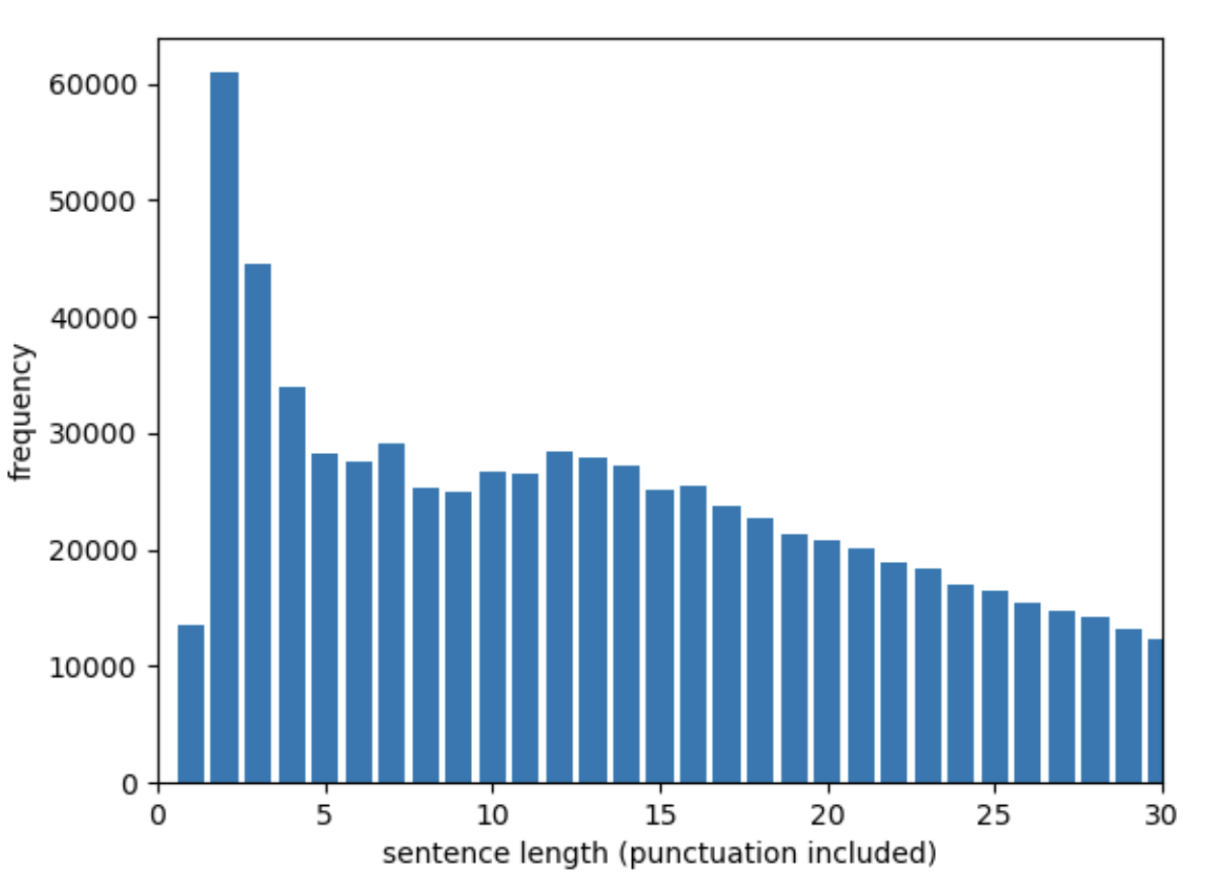}
        \caption{Cantonese corpus}
        \label{fig:cantonese_len}
    \end{subfigure}
    \caption{Distribution of sentence length.}
    \label{fig:data_len}
\end{figure}
After pre-processing, there are 912,258 lines of monolingual Cantonese data and 16M lines of monolingual Mandarin data. In terms of domains, the Cantonese corpus has 70\% data from Wikipedia while the Mandarin corpus is 100\% Wikipedia. Figure \ref{fig:data_len} shows that the distribution of sentence length in Cantonese and Mandarin are broadly similar after pre-processing. 

\begin{figure*}[htp]
    \centering
    \includegraphics[width=7.5cm]{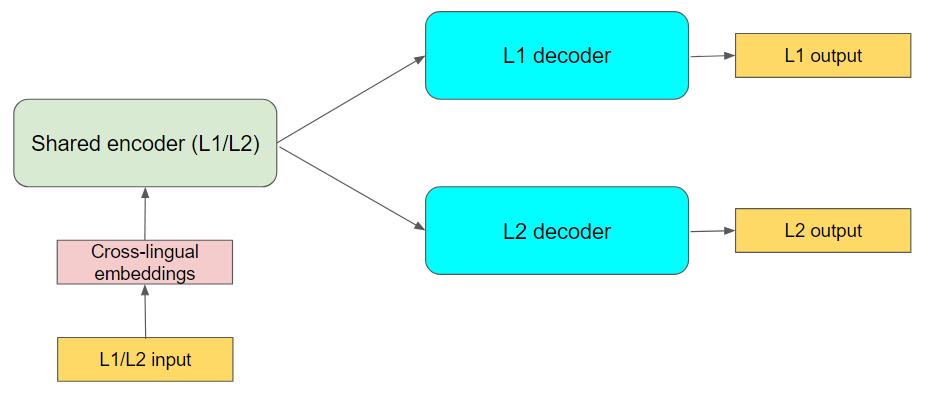}
    \caption{General architecture of the unsupervised machine translation systems in this experiment. A shared encoder maps sentences from L1/L2 to a common latent space, then a language-specific decoder reconstructs the encoded sentence back into its own language space. The model is trained by a denoising auto-encoding task and a back-translation task.}
    \label{fig:my_label}
\end{figure*}

\section{Methodology}

As shown in Figure \ref{fig:my_label}, we follow a standard unsupervised machine translation architecture with a shared encoder and language-specific decoders in our experiment. Models are trained on a denoising auto-encoding task and an on-the-fly back-translation task. To have an overall study of how different setups affect the model performance, we make three sets of comparisons:

\begin{enumerate}
    \item Model architectures.
    \item Cross-lingual embeddings.
    \item Tokenization methods.
\end{enumerate}

\subsection{Model Architectures}
In this experiment, we compare an RNN-based attention model and a transformer model.
\begin{itemize}
  \item RNN-based model: We adopt the architecture from \cite{artetxe2017unsupervised}: Both encoder and decoder have 2-layer bidirectional GRU \cite{cho2014properties}, Luong’s attention \cite{luong2015effective} is applied to align the source sentence and translation. Input sentences are converted to 512-dimensional cross-lingual embeddings. Considering the relatively lower capacity, the cross-lingual embeddings are fixed during training.
  \item Transformer model: Following \cite{lample2018phrase}, we use 4-layer encoder and decoder with 3-layer sharing parameters for both Cantonese and Mandarin sides. When generating translations, the decoder starts with a language-specific <BOS> token, specifying the language it is operating with. The embedding matrices are trainable during the training process.

\end{itemize}

\subsection{Cross-lingual Embeddings}
Cross-lingual embeddings can be learned in various different ways. In our experiments we compare the following three approaches:
\begin{itemize}
  \item Mapping: It has been extensively studied how to map monolingual word embeddings into a cross-lingual space.\cite{mikolov2013exploiting, artetxe2016learning, artetxe2017learning, artetxe2018generalizing, artetxe2018robust, conneau2017word}  In this project, we use Vecmap \footnote{https://github.com/artetxem/vecmap} by Artexte to obtain cross-lingual embeddings from monolingual ones. In particular, we adopt the “identical” setting, where the shared vocabulary in two languages can be used as anchors to learn the mapping. This approach is applied to RNN-based models.
  \item Learning from concatenated data: Another setup is to learn embeddings on the concatenation of source and target corpora in a monolingual way. As embeddings are learned in the context of both languages, the resultant embeddings can be seen as cross-lingual. This approach is applied on both RNN-based models and transformer models.
  \item Pivot-private embeddings: We also experiment with 512-dimensional pivot-private embeddings which consists of a 256-dimensional cross-lingual embedding learned on the concatenated dataset and a 256-dimensional private embedding, which is learned on two monolingual datasets separately. This approach is assumed to be able to capture the commonality between both languages and preserve language-specific characteristics as well \cite{wan2020unsupervised}. We adopt this approach on transformer models.

\end{itemize}

\subsection{Tokenization Methods}

We are also interested whether byte-pair encoding helps training Cantonese-Mandarin translation systems, so we compare it to a character-level tokenization method.

\begin{itemize}
    \item Word-level tokenization: As a baseline, we do no further tokenization on the collected data which is separated by words using Jieba and PyCantonese. In this setting, a total number of 80K/1M unique words are present in the Cantonese/Mandarin corpora respectively.
    \item Character-level tokenization: Since Mandarin and Cantonese are both analytic languages, character-level tokenization is a valid option to tokenize sentences. This results in 8K/14K unique tokens in Cantonese/Mandarin training data respectively.
    \item Byte-pair encoding: We also use byte-pair encoding to obtain a vocabulary of 50K sub-words on word-tokenized datasets. The embeddings of sub-words are learned using methods described above.
\end{itemize}

\section{Experiments and Results}

In this section, we describe the experiments we conducted and the results of both automatic and human evaluation. Our code and relevant repositories are publicly available online \footnote{https://github.com/meganndare/cantonese-nlp}.

\begin{table*}[t]
\begin{threeparttable}
\begin{tabular}{@{} *3l @{}}    
 \toprule
\emph{Model Name} & \emph{Can>Man Char BLEU } & \emph{Man>Can Char BLEU}  \\
 \midrule
Baseline (Character Conversion) Model & 13.3 & 13.2\\ 
 \midrule
RNN (Word Tok + Vecmap Embed) & 13.1 & 14.9\\ 
RNN (Char Tok + Vecmap Embed) & 19.8 & 22.5\\ 
RNN (Char Tok + Concat Embed) & 19.4 & 20.3\\
RNN (BPE Tok learned separately + Vecmap Embed) & 18.0 & 18.8\\ 
RNN (BPE Tok learned jointly + Vecmap Embed) & 19.3 & 19.5\\ 
 \midrule
RNN (Balanced Dataset + Word Tok + Vecmap Embed) & 6.2 & 11.5\\ 
RNN (Balanced Dataset + Char Tok + Vecmap Embed) & 17.1 & 20.4\\
 \midrule
Transformer (Char Tok + Concat Embed)** & 24.4 & 25.1\\ 
Transformer (Char Tok + Pivot-Private Embed) & 21.2 & 20.5\\ 
Transformer (BPE Tok learned jointly + Concat Embed) & 20.2 & 17.4\\
 \bottomrule
 \hline
\end{tabular}
\begin{tablenotes}
\item{Table 2: Overview of all automatic evaluation results. All BLEU (Bilingual Evaluation Understudy) metric scores are calculated at the character-level. Best-performing model indicated by **.}
\end{tablenotes}
\end{threeparttable}
\end{table*}

\subsection{Task Setup}
\subsubsection{Baseline Model}
Due to the large overlap in vocabulary between Mandarin and Cantonese and the lack of complicated morphology in both languages, for our baseline model we take advantage of these characteristics by evaluating Mandarin sentences as if they were a translation into Cantonese, and visa-versa. This method is carried out by simply converting both Mandarin and Cantonese evaluation datasets to the same character set using OpenCC \footnote{https://github.com/BYVoid/OpenCC} (our experiments used the \emph{Traditional Chinese (Hong Kong variant)} character set) and evaluating the BLEU score directly.

\subsubsection{RNN-based Experiments}
In order to improve upon the baseline model performance, we train several models using Artetxe's RNN+Attention-based architecture for unsupervised machine translation \footnote{https://github.com/artetxem/undreamt}. The primary objective, aside from improving BLEU scores over the baseline, is to identify which settings (e.g. tokenization scheme and embedding training method) lead to the best model performance. As detailed in the methodology section we experiment with word, character, and byte-pair encoding (BPE) tokenization, as well as cross-lingual embeddings obtained by learning a mapping into cross-lingual space, and by concatenation and training a skip-gram model. Additionally, for the BPE-tokenized models we have experimented with learning the BPE tokens separately for each language, or jointly.

\subsubsection{Balanced Dataset Experiments}
One characteristic of our full training dataset is that it is imbalanced (1 million Cantonese sentences versus 16 million Mandarin sentences). This is due to the abundance of Mandarin text data and the scarcity of Cantonese text data available. As a result, we were curious to understand whether having an imbalanced dataset negatively affects our training results. To this end we conducted an experiment using what we refer to as our 'Balanced Dataset'. To create the set, Mandarin sentences are chosen at random to be removed from the training set until a downsampled version of approximately the same size as the Cantonese training set was obtained, that also preserves the sentence length distribution of the original Mandarin training set. We then compare the performance of models trained using the balanced dataset to those trained using the full set, utilizing some simple baseline settings for comparison, namely word and character-tokenized models.

\subsubsection{Transformer Experiments}
Guided by advancements in neural network model architectures over the past several years, we are interested in how using a transformer architecture would impact our results. For the transformer experiments we leveraged Facebook Research's Unsupervised Neural Machine Translation Model \footnote{https://github.com/facebookresearch/UnsupervisedMT} for training. Using the results from our RNN-based models, we primarily focused on character and BPE tokenization schemes, and have also experimented with a more complex cross-lingual embedding type called pivot-private embeddings. Due to differences in implementation between the RNN and Transformer-based models, we were unable to train Vecmap embeddings for this set of experiments.

\subsection{Results}

\subsubsection{Automatic Evaluation}
\paragraph{Model Architectures}
The first metric that our study sought to investigate was the varying performances of Mandarin-Cantonese unsupervised machine translation based on the underlying neural network architecture, namely an RNN-based architecture versus a Transformer architecture. We observed that the transformer model led to higher BLEU scores when other factors are constant. This can be observed in the \emph{RNN (Char Tok + Concat Embed)} versus \emph{Transformer (Char Tok + Concat Embed)} models, where Cantonese-to-Mandarin translation yielded 19.4 versus 24.4, respectively; and Mandarin-to-Cantonese yielded 20.3 versus 25.1, respectively. In fact, our highest performing model from the study was trained on a Transformer architecture.

\paragraph{Cross-lingual Embeddings}
The study also makes comparisons between different types of cross-lingual embeddings. Of primary interest are training monolingual embeddings and mapping them to a shared cross-lingual space using Vecmap (as detailed in the Methodology section), and learning embeddings from the concatenated data. In a comparison between \emph{RNN (Char Tok + Vecmap Embed)} and \emph{RNN (Char Tok + Concat Embed)} models, we can see that the mapping-based cross-lingual embeddings have outperformed the concatenation-based technique, yielding a Cantonese-to-Mandarin BLEU of 19.8 and 19.4, respectively; and a Mandarin-to-Cantonese BLEU of 22.5 and 20.3, respectively.

In addition to mapping-based and concatenation-based cross-lingual embeddings, we also had time to run one experiment on pivot-private embeddings (as detailed in the Methodology section). By comparing the \emph{Transformer (Char Tok + Concat Embed)} and \emph{Transformer (Char Tok + Pivot-Private Embed)} models, we observe that concatenation-based embeddings outperform pivot-private embeddings, with a Cantonese-to-Mandarin BLEU of 24.4 versus 21.2, and a Mandarin-to-Cantonese BLEU of 25.1 to 20.5, respectively.

\paragraph{Tokenization Methods}
Our study additionally makes a comparison between different types of tokenization methods: word, character, and BPE-tokenized models. Word-tokenization always performs the worst, in all cases aside from one (see \emph{RNN (Word Tok + Vecmap Embed)} Mandarin-to-Cantonese results in Table 2), models trained with word-tokenized training data did not outperform even the \emph{Baseline (Character Conversion) Model} in which no neural network was trained.

While BPE-tokenized data tends to perform very well for languages with an alphabet system, such as French or English, we did not observe a such a strong result in the models trained using BPE-tokenized data for the Mandarin-Cantonese language pair. We experimented by learning BPE token vocabularies both separately and jointly, observing a slight performance improvement when learned jointly. However, neither BPE setting could outperform our character-tokenized models (see Table 2 for two results that lead to this conclusion: \emph{RNN (Char Tok + Vecmap Embed)} versus \emph{RNN (BPE Tok learned jointly + Vecmap Embed)}, as well as \emph{Transformer (Char Tok + Concat Embed)} versus \emph{Transformer (BPE Tok learned jointly + Concat Embed)}).

\paragraph{Balanced Dataset}
We conclude that neither word nor character-tokenized models trained on the balanced dataset outperformed models trained using the full training dataset. Thus, it is advantageous to use as much data as possible for model training, even if the two languages have an uneven amount of sentences.

\subsubsection{Human Evaluation}
We conduct human evaluation on the \emph{Transformer (Char Tok + Concat Embed)} model output in order to assess the extent to which our translation system would be useful to Cantonese and Mandarin speakers respectively. Considering that Cantonese speakers can understand Standard Chinese, a translation system from Mandarin to Cantonese should aim for localisation and fluency in Cantonese, while not losing the original meaning of the sentence. On the other hand, the primary purpose of a Cantonese-to-Mandarin translation system is to facilitate Cantonese comprehension for Mandarin speakers. For these diverging purposes in our translation directions, we manually evaluate each translation direction with separate criteria, which is explained in the following sections. 

\paragraph{Procedure} 100 lines from the test set are selected for evaluation, identical for both translation directions. One native speaker of each target language evaluates for that direction only (i.e. Cantonese speaker evaluates Mandarin to Cantonese sentences, and visa-versa). During evaluation, the evaluator has access to the original input and the target output. The evaluation decision is binary for both criteria, the evaluator can only choose either YES or NO. In the example sentences below, Mandarin features are highlighted in orange, Cantonese features are highlighted in teal and ungrammatical features are highlighted in red. 

\paragraph{Cantonese to Mandarin} System outputs are evaluated against the criteria concerning whether the output helps Mandarin speakers understand Cantonese text. 34\% were found helpful for understanding Cantonese text, 61\% were found not helpful, 5\% sentences are discarded because the original text in Cantonese is already perfectly comprehensible for Mandarin speaker.

\paragraph{Mandarin to Cantonese} System outputs are evaluated against the criteria \textit{``Does the system output contribute to Cantonese fluency / localisation?"}. It is found to be the case for 47\% of the sentences, false for 52\% of the sentences with 1\%sentences discarded since the input and target were identical. 

(1)-(4) are examples of the system output for the Mandarin to Cantonese direction. In (1), the output is evaluated as helpful even though it has not completely transformed all Mandarin features into Cantonese ones, however, the components with the highest semantic value (拍拖 \textit{dating} and 散 \textit{break up}) are in Cantonese where it was originally in Mandarin. Compared to (3), where the output still retains mostly Mandarin and has no Cantonese features. Comparing (2) and (4), they both have some grammatical errors (in red), but the impact of such error in (2) is less significant to the overall meaning of the sentence, while in (4) the overall sentence is incomprehensible. 

\begin{displayquote}
\underline{Examples of output that is helpful:} \\
(1)\\
Mandarin reference (source):\\
身邊有兩位好朋友，\textcolor{orange}{交往了}三年，就\textcolor{orange}{那樣}分手\textcolor{orange}{了}。\\
Cantonese reference (target): \\
身邊有兩位好友，\textcolor{teal}{拍咗}三年\textcolor{teal}{拖}，\textcolor{teal}{就噉散咗}。\\
System output: \\
身邊有兩位好友，\textcolor{teal}{拍}\textcolor{orange}{了}三年\textcolor{teal}{拖}，就\textcolor{orange}{這樣}\textcolor{teal}{散}\textcolor{orange}{了}。\\
Sentence meaning: \textit{I have two friends who had been dating for three years, and they broke up just like that.}\\

(2)\\
Mandarin reference (source):\\
\textcolor{orange}{別這麼\begin{CJK*}{UTF8}{gbsn}犟\end{CJK*}}，快\textcolor{orange}{點}向媽認錯。\\
Cantonese reference (target): \\
\textcolor{teal}{咪咁硬頸}，快\textcolor{teal}{啲同亞}媽認錯。\\
System output: \\
\textcolor{red}{否“}\textcolor{teal}{硬頸}，快\textcolor{orange}{些和}\textcolor{teal}{亞}媽認錯 。\\
Sentence meaning: \textit{Don't be so stubborn, apologize to your mother at once.}\\

\underline{Examples of output that is not helpful:} \\
(3)\\
Mandarin reference (source):\\
\textcolor{orange}{別小看他}，\textcolor{orange}{他}已經有\textcolor{orange}{了}三項發明。\\
Cantonese reference (target): \\
\textcolor{teal}{咪睇小佢}，\textcolor{teal}{佢}已經有\textcolor{teal}{咗}三項發明。\\
System output: \\
\textcolor{red}{否}\textcolor{orange}{看小她}，\textcolor{orange}{她}已經\textcolor{orange}{有了}三項發明。\\
Sentence meaning: \textit{Don't underestimate him, he already has three inventions.}\\

(4)\\
Mandarin reference (source):\\
\textcolor{orange}{給}海關沒收\textcolor{orange}{了那些東西}。\\
Cantonese reference (target): \\
\textcolor{teal}{畀}海關\textcolor{teal}{執咗嗰啲嘢}。\\
System output: \\
\textcolor{orange}{給}海關\textcolor{teal}{執}\textcolor{orange}{了}\textcolor{red}{那麼}。\\
Sentence meaning: \textit{The things that were confiscated by customs.}\\
\end{displayquote}

\section{Discussion}

Our Mandarin-Cantonese machine translation project displays the differences between two tokenization methods (character-level and byte pair encoding), with an outcome different than expected regarding byte pair encoding. A possible reason for this may be that such a big vocabulary size can lead to worse embeddings, taking into account the size of our corpus.

One of our approaches was down-sampling the full dataset into a balanced one, from which we expected a higher BLEU score compared to when using the full dataset. However, this had the opposite effect on the BLEU score and it ended up being lower than in the previous occasions. This is perhaps due to the fact that 1 million sentences is just simply not enough data for a machine to become 'fluent' in a language.

As further work, we propose that this project can be extended by combining out best architecture, best tokenization and best embedding training method (transformer + character + mapping), by developing a cross-lingual mapping for embeddings that is compatible with a transformer network in order to confirm whether it does lead to higher results. 

In addition, other options worth exploring would be the grammatical similarity between Cantonese and Mandarin and developing an statistical machine translation model.

\section{Summary and conclusion}
The aim of implementing a Cantonese-Mandarin MT-model was accomplished by: 
        \begin{itemize}
            \item Creating a large-scale corpus out of several online sources such as Wikipedia, scraped Instagram comments, YouTube subtitles and restaurant reviews.
            \item Implementing and training several Cantonese-Mandarin translation models while studying the effects of different tokenization strategies, such as character-level and byte-pair encoding. While BPE was expected to outperform character-level tokenization, this was not the case in our experiments. 
        \end{itemize}
The outcomes of this project showed that overall, in 61\% of the cases, the outcome translation was not useful to help Mandarin speakers understand Cantonese text. As far as what fluency concerns, in 52 out of 100 cases, the system's output did not show any contribution.

Further work and research is essential in order to reach good percentages of performance and fluency in such a machine translation model. This project has contributed a large Cantonese dataset that was not available before as it is now.

We hope that with this project we moved one step forward into a direction that has been studied for some years now, contributing to further developments and advancement.

\bibliography{anthology,custom}
\bibliographystyle{acl_natbib}

\end{CJK*}
\end{document}